\pdfoutput=1

\documentclass[11pt]{article}

\usepackage{acl}

\usepackage{times}
\usepackage{latexsym}

\usepackage[T1]{fontenc}

\usepackage[utf8]{inputenc}

\usepackage{microtype}

\usepackage{graphicx}
\usepackage{float}
\usepackage{subcaption}

\usepackage{amsmath}
\usepackage{amssymb}

\usepackage{algorithm}
\usepackage{algpseudocode}

\usepackage{multirow}

\usepackage{array}

\title{Codec at SemEval-2022 Task 5: Multi-Modal Multi-Transformer Misogynous Meme Classification Framework}

\author{Ahmed Mahran \\
  Codec AI \\
  Alexandria, Egypt \\
  \texttt{ahmed@codec.ai} \\\And
  Carlo Alessandro Borella \\
  Codec AI \\
  WC1N 2EB, London, UK \\
  \texttt{carlo@codec.ai} \\\And
  Konstantinos Perifanos \\
  Codec AI \\
  WC1N 2EB, London, UK \\
  \texttt{kostas@codec.ai} \\}

\begin{document}
\maketitle
\begin{abstract}

In this paper we describe our work  towards building a generic framework for both multi-modal embedding and multi-label binary classification tasks, while participating in task 5 (Multimedia Automatic Misogyny Identification) of SemEval 2022 competition.

Since pretraining deep models from scratch is a resource and data hungry task, our approach is based on three main strategies. We combine different state-of-the-art architectures  to capture a wide spectrum of semantic signals from the multi-modal input. We employ a multi-task learning scheme to be able to use multiple datasets from the same knowledge domain to help increase the model's performance. We also use multiple objectives to regularize and fine tune different system components.

%
\end{abstract}

\begin{figure*}[ht]
    \begin{subfigure}[t]{0.5\textwidth}
        \centering
        \includegraphics[width=\linewidth]{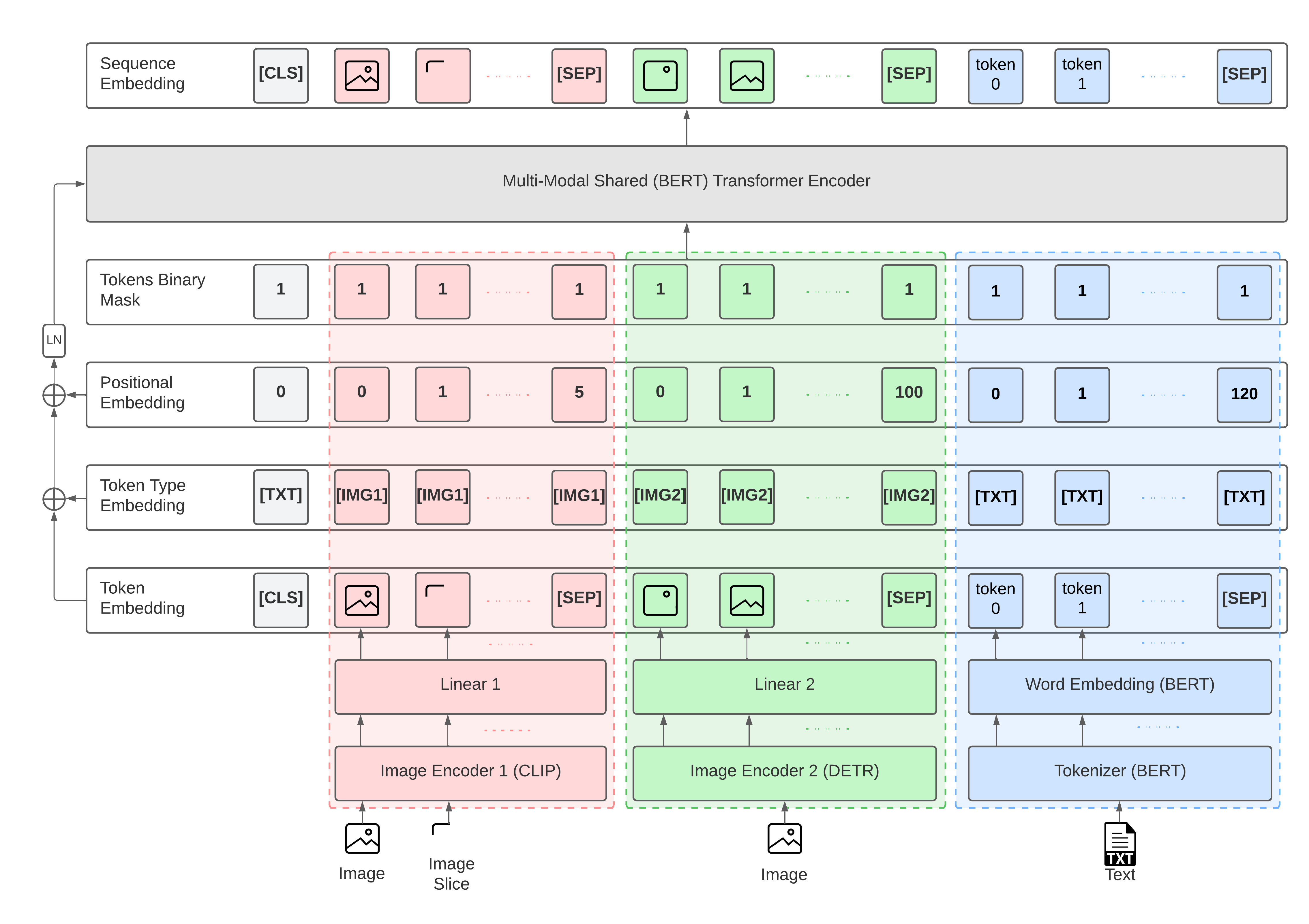}
        \caption{Multi-modal shared transformer encoder}
        \label{fig:shared_transformer}
    \end{subfigure}
    ~
    \begin{subfigure}[t]{0.5\textwidth}
        \centering
        \includegraphics[width=\linewidth]{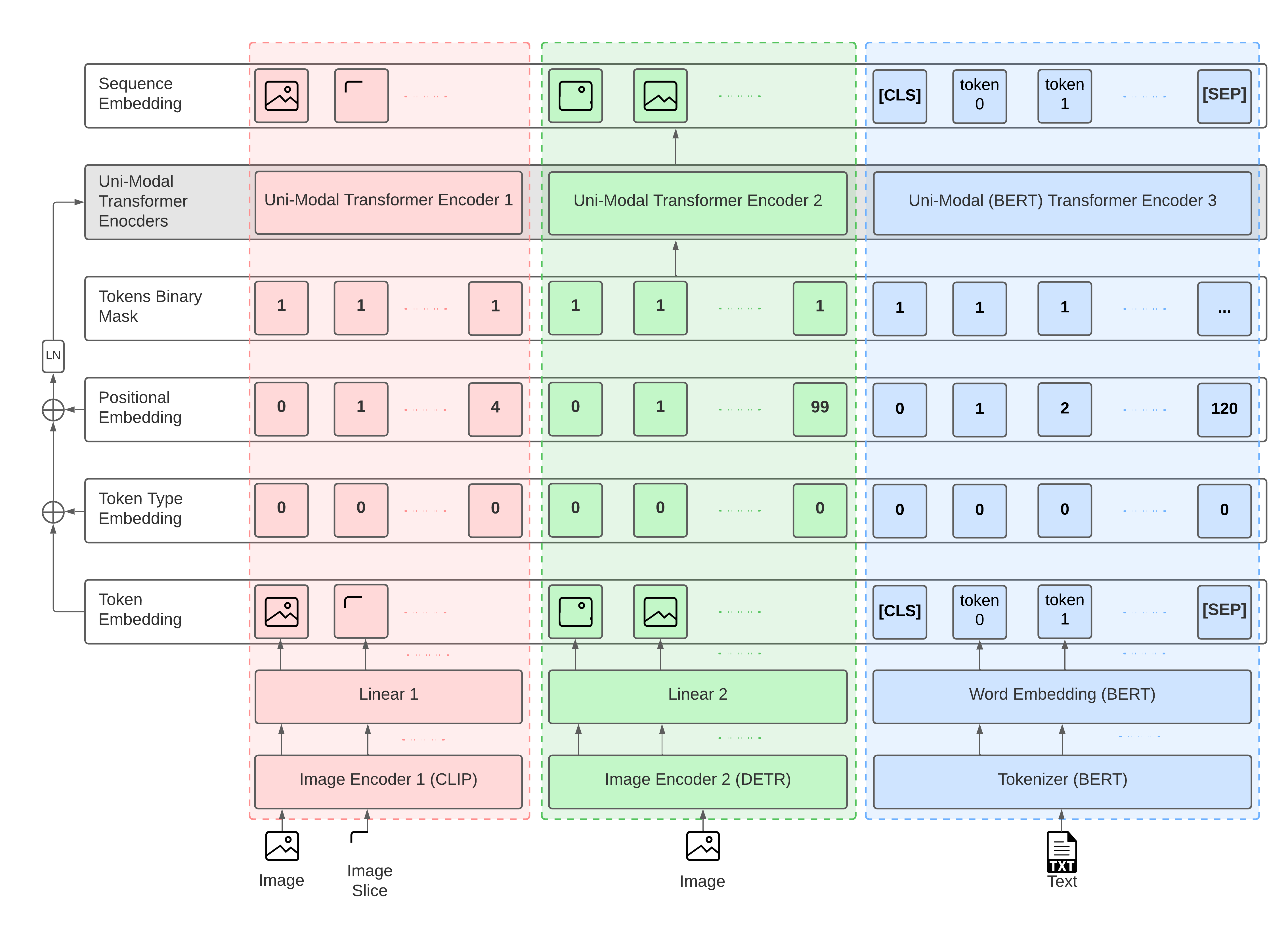}
        \caption{Uni-modal multi transformer encoders}
        \label{fig:multi_transformers}
    \end{subfigure}
    \caption{Sequence embedding}
    \label{fig:transformer}
\end{figure*}

\section{Introduction}

In this paper, we present the system that we have built to participate in SemEval 2022 task 5 \cite{task5}, Multimedia Automatic Misogyny Identification (MAMI) challenge. The task is targeted at identification of misogynous memes by basically using the meme's image and pre-extracted English text content as input sources. The task is divided into two main sub-tasks: Sub-task A is a binary classification task where a meme should be categorized either as misogynous or not misogynous, Sub-task B is a multi-label binary classification task, where the type of misogyny should be recognized among the potential overlapping categories: stereotype, shaming, objectification and violence. Generally, meme classification is a challenging task as memes are multi-modal, rely heavily on implicit knowledge, and are subject to human misinterpretation especially among different backgrounds and cultures.

We have used transformer \cite{vaswani2017attention} based architectures and took a transformer based approach to combine them. Transformer based architectures are achieving state-of-the-art performance for Natural Language Processing and Computer Vision related tasks. There are architectures for language, vision, and language and vision combined. There are also architectures for other modalities however in MAMI's scope, we are interested in images and text only. 

Pretraining a deep neural network and a transformer based architecture from scratch is a data hungry and computation resources demanding task. Especially in a multi-modal domain where input can have multiple image and text modalities. The literature is rich in a variety of architectures which achieve competitive performance in different tasks for different modalities. In our work, we took an approach towards building a framework that would allow us to combine different pretrained architectures. With relatively few epochs and using relatively less compute resources, our goal was to build and train a classifier framework that could harness the power of pretrained architectures as backbones, using relatively limited resources: no multiple GPUs for training and constrain the cost to be relatively small, in the range of hundred dollars in total.

We have assumed that using as many different backbone architectures which are trained on different tasks for different modalities can allow us to capture a wide spectrum of semantic signals from the input modalities. Then we just need to build a classifier to learn the relationship between these signals and the target classes.

We have also found and discuss below that there are available text and image datasets for hate speech, sexism and hateful memes which sound related to this task and could be used to augment MAMI's dataset. Eventually, our goal was to combine different datasets in a multi-task classifier.

In order to guide the model during training towards the main objective, which is to minimize the classification loss, we have employed different auxiliary objectives. Breaking down the whole architecture into components, each component has a sub-target. We have assumed that if we could assure that each component performs well on the sub-target, then the whole system would perform better on the main target. A component producing a clear signal can facilitate the learning of the downstream dependent components. This could increase model performance in terms of either accuracy or convergence speed. So, if we could formulate an objective function for each component, we can linearly combine them with the main objective function. This helps fine tuning and regularizing the components of the system.

We have built a generic classification framework and applied it to both sub-tasks A and B. During the evaluation phase of the competition, we have achieved, for sub-task A, a macro-average F1-Measure of \(0.715\), and for sub-task B, a weighted-average F1-Measure of \(0.698\). During post-evaluation phase we have achieved higher score for sub-task A of \(0.761\). Our code is available at \url{https://github.com/ahmed-mahran/MAMI2022}.

\section{Background}

\textbf{Datasets:} Besides MAMI dataset, there are many datasets that could be used to train our model. The input could be uni-modal as text only or image only, or bi-modal as pairs of image and text. \cite{vidgen2020directions} reviews \(63\) publicly available training datasets and they have published a dataset catalogue on a dedicated website \footnote{\href{https://hatespeechdata.com/}{hatespeechdata.com}}. We have used the hateful meme dataset created by Facebook AI \cite{kiela2020hateful} which consists of \(10K\) memes labeled hateful or not.

\textbf{Multi-modal frameworks:} MMBT \cite{kiela2019supervised} concatenates, into a single sequence, linear projection of ResNet \cite{he2016deep} output for image pooled to \(N\) different vectors, with BERT \cite{devlin2018bert} tokens embeddings for text. The sequence is fed into a transformer encoder, they call it a bi-transformer, after adding positional embeddings and segment embeddings to distinguish which part is image and which part is text. The architecture is generic enough to use different image encoders. As a variant of how we combine signals from image and text, we extend the MMBT architecture to combine more than two encoders however that wasn't our top performing variant. MMCA \cite{wei2020multi} combines Faster R-CNN \cite{ren2015faster} with BERT to compute two embedding types for each modality: self-attention embedding to capture intra-modality interactions, and cross-attention embedding to capture both intra- and inter-modality interactions.

\begin{figure*}[ht]
    \centering
    \includegraphics[width=\textwidth]{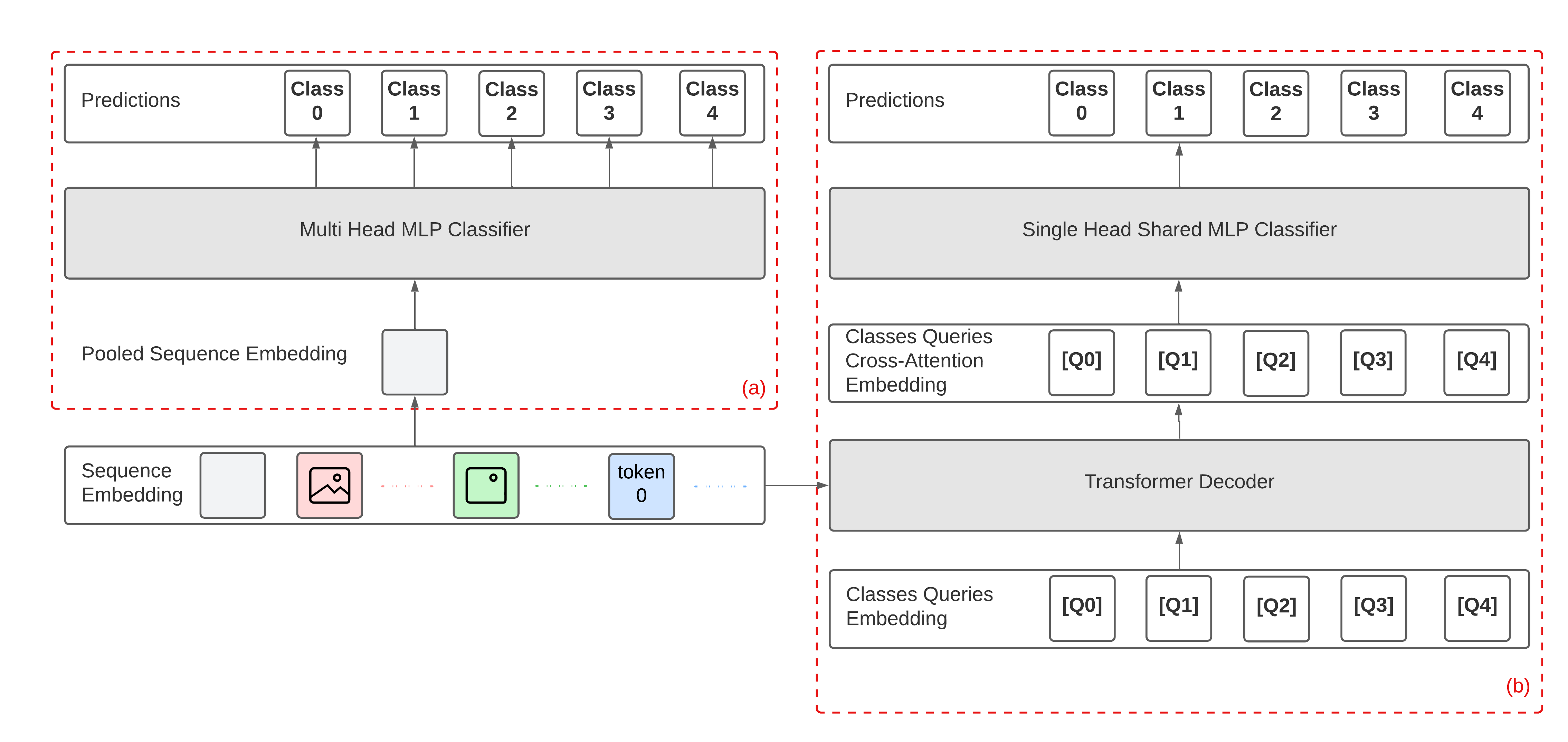}
    \caption{Classifiers: (a) using pooled sequence embedding, (b) using the whole non-pooled sequence embedding.}
    \label{fig:classifier}
\end{figure*}

\section{System overview}
\subsection{Tokens embedding}\label{sec:tokens_embedding}
At this stage, we encode each input modality into a sequence of vectors in a unified dimension space. We also generate a binary mask vector with length equal to the sequence's length to indicate which part of the sequence the model should consider. This produces the output at "Token Embedding" and "Tokens Binary Mask" layers illustrated in figure \ref{fig:transformer}. We can use different encoders per modality and generate different sequence types to capture as many semantic signals as possible. For our setup, we have tested CLIP \cite{radford2021learning} and DETR \cite{carion2020end} encoders for image modality and BERT \cite{devlin2018bert} encoder for text modality.

\textbf{CLIP image embedding:} We split the image into \(4\) equal size patches (\(2\times2\)) then we use CLIP \cite{radford2021learning} to encode the whole image along with its \(4\) slices into a sequence of \(5\) vectors. Then we project each vector into the model hidden dimension space. In our experiments, we have used CLIP model named "RN50x4" (which uses as backbone ResNet-50 scaled up 4x using the EfficientNet scaling rule \cite{tan2019efficientnet})\footnote{We have used OpenAI implemantion on \url{https://github.com/openai/CLIP}}.

\textbf{DETR image objects embedding:} We use DETR \cite{carion2020end} \footnote{We have used \href{https://huggingface.co/docs/transformers/v4.18.0/en/model_doc/detr}{ Huggingface Transformers implementation of DETR}} to encode the image into a sequence of \(100\) image objects representations. DETR's transformer decoder produces a sequence of \(100\) possible object boxes representations that we use as objects embeddings. However, not all of the objects are real objects as DETR can produce a no-object prediction. So, we use DETR's classifier which is trained on the object detection task to generate masks for no-objects. For each box from the \(100\), the classifier produces \(92\) logits which correspond to \(92\) possible object labels. We take the softmax of the \(92\) logits and mask out the corresponding object box if the label with the highest softmax probability is the no-object label. We fallback to another masking strategy if all the \(100\) boxes are masked out; we ignore the logit of the no-object label and then take the softmax of the rest labels to select \(4\) out of \(100\) boxes with highest softmax probabilities. Similarly with CLIP output, we project each vector into the model hidden dimension space.

\textbf{BERT text embedding:} We use BERT \cite{devlin2018bert} \footnote{We have used \href{https://huggingface.co/docs/transformers/v4.18.0/en/model_doc/bert}{ Huggingface Transformers implementation of BERT} with weights from \href{https://huggingface.co/Hate-speech-CNERG/bert-base-uncased-hatexplain}{"Hate-speech-CNERG/bert-base-uncased-hatexplain"}} pre-trained on hate speech \cite{mathew2020hatexplain} to tokenize input text and generate tokens embeddings (for a maximum of \(120\) tokens). We don't project BERT's embeddings as we use its output space as the model's hidden dimension space (which has length of \(768\) dimensions).

\subsection{Sequence embedding}
At this stage, the model generates one combined sequence of vectors using tokens embedding from the tokens' embedding stage. This is the output in the "Sequence Embedding" layer illustrated in figure \ref{fig:transformer}. For the token embedding output, we add token type embedding and positional embeddings that encode the position of each token in the corresponding input sequence per type then we apply a layer normalization \cite{ba2016layer}. Along with input masks, the layer normalized sum of embeddings is fed to a multi-layer transformer encoder stage to produce the final sequence embedding at the "Sequence Embedding" layer in figure \ref{fig:transformer}. The output sequence has the same number and dimensionality of the input tokens. We have two variants for the transformer encoder stage:

\textbf{Shared transformer encoder:} The general architecture of the shared transformer encoder variant is illustrated in Figure \ref{fig:shared_transformer}. What distinguishes this variant is that we use a shared multi-layer transformer encoder to capture the intra- and inter-modality interactions. Because of this, for each input embedding type, we add token type embedding which is the same for each input encoder to distinguish which tokens are from which input encoder. In our setup, we have three token types; that is one for each of: CLIP, DETR and BERT encoders. We also append a special [SEP] token at the end of each sequence type and we prepend to the whole combined sequence a global and special [CLS] token. We use the pre-trained BERT transformer encoder as the shared transformer encoder.

\textbf{Multiple transformer encoders:} The general architecture of this variant is illustrated in Figure \ref{fig:multi_transformers}. The difference here is that instead of using one shared transformer encoder, we use a multi-layer transformer encoder per token type. We still add a token type embedding such that each transformer encoder learns its own token type parameters. We think we can remove this step however we have not tested this. Also, in this variant we don't need to add the extra [SEP] token per type and the global [CLS] token however we add a local [CLS] and [SEP] tokens for BERT only. For CLIP and DETR, we use PyTorch's implementation of the transformer encoder which is described in \cite{vaswani2017attention} with \(8\) heads and \(6\) layers but for BERT we keep it as in \cite{devlin2018bert}. Then the final output sequence is just the concatenation of all sequences from each transformer encoder.

\subsection{Classification}
We have two modes of classification depending on the sequence length of the output of the transformer encoder stage. As shown in figure \ref{fig:classifier}, we either use the whole sequence of vectors or pool it to one vector.

\textbf{Multi-head MLP classifier:} We use the pooled sequence embedding as classifier input. In our setup, we have used the first [CLS] token in the shared transformer encoder variant. The classifier is a two feed forward linear layers with a GELU activation in between and a hidden size of \(768\). The final layer produces number of logits equals to number of classes and we apply a sigmoid activation to compute each binary label probability.

\textbf{Transformer decoder with single-head shared MLP classifier:} Here the whole sequence embedding along with the binary mask is fed into a multi-head transformer decoder as a source sequence. Then the target sequence is formed by a learnable target class query embedding for each class in the target classes. The transformer decoder learns how each class interacts with each input modality signal through the cross-attention mechanism between the source and target sequences. Moreover, the decoder learns the dependency among the target classes through the self-attention mechanism for the target sequence. The decoder output is then fed into a single-head MLP classifier that shares parameters for all classes such that \(MLP(q_i)\) is the logit of label \(i\) using the corresponding class query embedding, \(q_i\), from the decoder output. The MLP classifier has the same architecture as the previous one in terms of number of layers, type of activation and hidden size, and similarly as well we compute the binary label probability for each class. We use PyTorch's implementation of the transformer decoder which is described in \cite{vaswani2017attention} with \(8\) heads and \(6\) layers.

\subsection{Multi-task learning}
In order to use more training data from other but similar datasets, we have followed a multi-task learning approach. For each dataset \(\mathcal{D}\), there is a set of target labels \(\mathcal{L}\), we can define as many tasks as the sets of labels in the power set \(\mathcal{P}^{+}(\mathcal{L})\) (excluding the empty set) such that it is possible to use the same label in more than one task. Each task has a separate MLP classifier while all tasks across the datasets share the rest of the parameters including the learnable classes queries. During training, each mini-batch contains data from only one dataset and we compute the targets per task for all the tasks of the dataset.

\begin{algorithm}
\caption{Multi-task learning}\label{alg:multi_task}
\begin{algorithmic}
\State $datasets \gets \{D_1, D_2, ...\}$
\State $tasksPerD \gets \{(D_1, \{T_1, T_2, ...\}), ...\}$
\State $labelsPerT \gets \{(T_1, \{l_1, l_2, ...\})\}$
\For{$epoch \in epochs$}
    \For{$b \in mini\text{-}batches$}
        \State $dataset \gets \textbf{sample}\text{ }1\text{ }\textbf{from}\text{ }datasets$
        \State $tasks \gets tasksPerD[dataset]$
        \For{$task \in tasks$}
            \State \textbf{learn} $b,\text{ }task,\text{ }labelsPerT[task]$
         \EndFor
    \EndFor
\EndFor
\end{algorithmic}
\end{algorithm}

\subsection{Multi-objective}
For a training instance \(i\), we use \(x_i\) to refer to the input regardless from its actual representation, \(y_{i,c} \in \{0,1\}\) is the value of the target binary label \(c\), and \(\theta\) is the set of learnable parameters.

\textbf{Main objective}: We use a binary cross entropy to minimize the loss per label.
\begin{equation}
\begin{split}
\mathcal{L}_0(i,c) = \ y_{i,c} . \log p(c|x_i,\theta) \\
+ (1 - y_{i,c}) . \log(1 - p(c|x_i,\theta))
\end{split}
\end{equation}

\textbf{Token encoding projection alignment:} We linearly project modal encoding into the hidden model space (as described in section \ref{sec:tokens_embedding}). In order to preserve a similar structure of data points across spaces, we impose a cosine similarity constraint such that for any two input instances, \(x_i\) and \(x_j\), the similarity, \(s(., .)\), between their encoding, \(f(.)\), is the same as the similarity between the projection, \(g(.)\), of their encoding. We apply this to all pairs in each batch.
\begin{equation}
\begin{split}
\mathcal{L}_1(i,j) = \ |s(f(x_i), f(x_j)) \\
- s(g(f(x_i)), g(f(x_j)))|
\end{split}
\end{equation}

\textbf{Contrastive embedding loss:} This is intended at regularizing the embedding space of the MLP classifier to make instances of dissimilar labels more separable. For any two input instances, \(x_i\) and \(x_j\), we apply the embedding loss per class \(c\) on the input, \(h^{l-1}\), of each layer \(l\) of the MLP classifier.
\begin{equation}
\begin{split}
\mathcal{L}_2(i,j,c) &=\\
\frac{1}{2} \sum_{l}1- s_c(y_{i}, y_{j})s_h(h^{l-1}_{i,c},h^{l-1}_{j,c})
\end{split}
\end{equation}
\begin{equation}
\begin{split}
s_c(y_i, y_j) = (2y_{i,c}-1)(2y_{j,c}-1)
\end{split}
\end{equation}
\(s_c(., .) \in \{-1,1\}\) is the labels similarity for class \(c\) of two instances; \(-1\) indicates dissimilar labels while \(1\) indicates similar labels. \(s_h(., .) \in [-1,1]\) is the cosine similarity of layer input when comparing two instances. It is worth noting that in case of the multi-head MLP classifier, \(h^{l-1}_{i,c}\) is the same for all classes. We also apply this loss to all pairs in each batch. This loss encourages the transformer encoder in case of the multi-head MLP classifier, the decoder in case of the shared single head MLP classifier, as well as the hidden layers of the MLP classifier to produce embeddings with structures that capture labels similarity such that instances with the same label value get closer embeddings than instances with different label value.

The overall loss per batch and task given a dataset \(d\):
\begin{equation}
\begin{split}
\mathcal{L}_t^d = \frac{1}{N_iN_c^t}\sum_{i,c}\mathcal{L}_0(i,c) \\
+ \frac{1}{N_i(N_i-1)} \sum_{i \neq j} \mathcal{L}_1(i,j)\\
+ \frac{1}{N_i(N_i-1)N_c^t} \sum_{i \neq j,c} \mathcal{L}_2(i,j,c)
\end{split}
\end{equation}

The overall loss per batch for all tasks of the dataset is the average task loss:
\begin{equation}
\begin{split}
\mathcal{L}^d = \frac{1}{N_t^d}\sum_{t}\mathcal{L}_t^d
\end{split}
\end{equation}
\(N_i\) is the batch size, \(N_t^d\) is the number of tasks for dataset \(d\), and \(N_c^t\) is the number of classes for task \(t\).

\section{Experimental setup}

In addition to MAMI's dataset, we have used Facebook's hateful memes dataset. We have set the tasks configurations as shown in table \ref{tbl:tasks_config}.

\begin{table}[ht]
\centering
\small
\begin{tabular}{lll}
  \hline
 Dataset & Task & Labels \\ 
  \hline
  \hline
  \multirow{5}{3em}{MAMI} & MAMI & \{misogynous,\ shaming ,\ stereotype \\ 
& & ,\ objectification,\ violence\} \\
& Task\_A & \{misogynous\} \\
& Task\_B & \{shaming,\ stereotype \\ 
& & ,\ objectification,\ violence\} \\
   \hline
   FBHM & Hateful & \{hateful\} \\
   \hline
\end{tabular}
\caption{Tasks labels configurations per dataset. FBHM is short for Facebook Hateful Meme.}
\label{tbl:tasks_config}
\end{table} 
We have added the redundant task MAMI for the MAMI dataset to make sure that the decoder learns the dependency among all labels as Task\_B's labels depend on Task\_A's label we wanted to make sure that the model captures this dependency.

We have split both datasets into \(80\%\) train and \(20\%\) dev sets using stratified sampling. We have used the data provided at post-evaluation period of the competition as the test set. We have used a batch size of \(16\) and number of epochs as \(15\). We have used MADGRAD \cite{defazio2021adaptivity} for optimization. Learning rate was set to \(2 \times 10^{-4}\) and we used a learning rate linear scheduler with a warmup period \footnote{We used \href{https://huggingface.co/docs/transformers/v4.18.0/en/main_classes/optimizer_schedules\#transformers.get_linear_schedule_with_warmup}{get\_linear\_schedule\_with\_warmup} from Huggingface Transformers.}. Gradients are accumulated every \(20\) batches so there was a \(\text{total gradient accumulation steps}\) of: \(\text{total number of batches} / 20 \times \text{number of epochs}\). We have set the warmup period to: \(\text{total gradient accumulation steps} / 10\). For all parameters except biases and layer normalization weights, we have used a weight decay of \(5 \times 10^{-4}\). We clip gradients to overall norm of \(0.5\). Our implementation is PyTorch based and we have used HuggingFace Transfomers implementation for both BERT and DETR. We have run our experiments on Google Colab Pro plus using one Tesla P100 GPU. We didn't perform any special data preprocessing, just the requirements for each backbone. Also, to make experiments quicker, we didn't perform any fine tuning for any of the backbones and we pre-generated and stored each backbone output instead of re-evaluating the same data across epochs and experiments.

We used the official accuracy measures to score how the model performs on each task. For the single class tasks, we have used macro-average F1-Measure and we called it scoreA. In particular, for each class label (i.e. true and false) the corresponding F1-Measure will be computed, and the final score will be estimated as the arithmetic mean of the two F1-Measures. For the multi class tasks, we have used weighted-average F1-Measure and we called it scoreB. In particular, the F1-Measure will be computed for each label and then their average will be weighted by support, i.e. the number of true instances for each label.

\section{Results}
\subsection{Ablations}\label{sec:ablations}
We have the following system configurations \footnote{We give each item a short name in parenthesis to be able to refer to corresponding items in experiments results tables compactly.} which would result in different architecture variations.
\begin{description}
\item[Transformer encoder (Xformer Enc)] whether to use shared transformer (\textbf{Shared}) or multi transformers (\textbf{Multi}).
\item[Encoder output pooling (Pooling)] whether the whole sequence is pooled using [CLS] embedding (\textbf{[CLS]}), this means we use the multi-head MLP classifier and no decoder), or no pooling (\textbf{No}) or only text tokens are pooled using text's [CLS] token (\textbf{txt [CLS]}), this means we use the decoder with the shared single head MLP classifier.
\item[Token encoding projection alignment (Proj Align)] whether to enable it (\textbf{Yes}) or not (\textbf{No}).
\item[Contrastive embedding loss (Contrastive)] whether to enable it (\textbf{Yes}) or not (\textbf{No}).
\item[Multi-task learning (Multi-task)] whether to use Facebook's hateful meme dataset (\textbf{Yes}) or not (\textbf{No}).
\item[Image encoders (Backbones)] whether to use CLIP only (\textbf{CLIP}), DETR only (\textbf{DETR}), or both together (\textbf{CLIP and DETR}).
\end{description}
We plan experiments as tournament of rounds such that in each round we test subset of the configurations fixing the rest. Then we use the winning configuration values for subsequent rounds. For each experiment, we report the max score for Task\_A and Task\_B on all splits. Appendix \ref{sec:appendix} contains more details on scores distributions.

\subsubsection{Round 1}
At this round we compare transformer encoder and encoder output pooling methods. We perform four experiments with configurations and results summaized in table \ref{tbl:exp_round1}. The winner of this round is the multi-transformers encoders without output pooling architecture variant.
\begin{table}[ht]
\centering
\small
\begin{tabular}{l|llll}
    \hline
    Experiment & 00 & 01 & 02 & 03 \\ 
    \hline
    \hline
    Xformer Enc & Shared & Shared & Multi & Multi \\
    Pooling & [CLS] & No & No & txt [CLS] \\
    Proj Algn & \multicolumn{4}{c}{No} \\
    Contrastive & \multicolumn{4}{c}{No} \\
    Multi-task & \multicolumn{4}{c}{No} \\
    Backbones & \multicolumn{4}{c}{CLIP and DETR} \\
    \hline
    Score \\
    \hline
    Test - Task\_A & 0.6819 & 0.7226 & \bf{0.7436} & 0.7329 \\
    Test - Task\_B & 0.5886 & 0.6422 & \bf{0.6785} & 0.6772\\
    Dev - Task\_A & 0.8395 & 0.8355 & \bf{0.8564} & 0.8519\\
    Dev - Task\_B & 0.6650 & 0.6933 & \bf{0.7420} & 0.7310\\
    Train-Task\_A & \bf{0.9253} & 0.9121 & 0.9066 & 0.8998\\
    Train-Task\_B & 0.7006 & 0.7242 & \bf{0.7782} & 0.7647\\
    \hline
\end{tabular}
\caption{Experiments configurations for round 1 and corresponding results.}
\label{tbl:exp_round1}
\end{table}

\subsubsection{Round 2}
At this round, we test the significance of the multi-objective approach. The configurations and corresponding results are summarized in table \ref{tbl:exp_round2}. Token encoding projection alignment makes improvements on both tasks on the test split when enabled alone. Also, contrastive embedding loss seems to slightly improve Task\_B. After performing statistical tests and comparing distributions of the scores, we pick experiment 10 as the winner variant.

\begin{table}[ht]
\centering
\small
\begin{tabular}{l|l|lll}
    \hline
    Experiment & 02 & 10 & 12 & 13 \\ 
    \hline
    \hline
    Xformer Enc & \multicolumn{4}{c}{Multi} \\
    Pooling & \multicolumn{4}{c}{No} \\
    Proj Algn & No & Yes & No & Yes \\
    Contrastive & No & No & Yes & Yes \\
    Multi-task & \multicolumn{4}{c}{No} \\
    Backbones & \multicolumn{4}{c}{CLIP and DETR} \\
    \hline
    Score \\
    \hline
    Test - Task\_A & 0.7436 & \bf{0.7504} & 0.7358 & 0.7467\\
    Test - Task\_B & 0.6785 & 0.6798 & \bf{0.6823} & 0.6777\\
    Dev - Task\_A & \bf{0.8564} & 0.8535 & 0.8515 & 0.8535\\
    Dev - Task\_B & \bf{0.7420} & 0.7387 & 0.7371 & 0.7313\\
    Train-Task\_A & 0.9066 & 0.8983 & \bf{0.9090} & 0.8988\\
    Train-Task\_B & \bf{0.7782} & 0.7679 & 0.7676 & 0.7573\\
    \hline
\end{tabular}
\caption{Experiments configurations for round 2 and corresponding results compared to best configurations from round 1.}
\label{tbl:exp_round2}
\end{table}

\subsubsection{Round 3}\label{sec:abl_round3}
At this round, we test the significance of the image encoders backbones, namely: CLIP and DETR. The configurations and corresponding results are summarized in table \ref{tbl:exp_round3}. It seems that our use of DETR was incompetent to CLIP.

\begin{table}[ht]
\centering
\small
\begin{tabular}{l|l|ll}
    \hline
    Experiment & 10 & 20 & 21 \\ 
    \hline
    \hline
    Xformer Enc & \multicolumn{3}{c}{Multi} \\
    Pooling & \multicolumn{3}{c}{No} \\
    Proj Algn & \multicolumn{3}{c}{Yes} \\
    Contrastive & \multicolumn{3}{c}{No} \\
    Multi-task & \multicolumn{3}{c}{No} \\
    Backbones & CLIP and DETR & CLIP & DETR \\
    \hline
    Score \\
    \hline
    Test - Task\_A & \bf{0.7504} & 0.7426 & 0.7010\\
    Test - Task\_B & 0.6798 & \bf{0.6865} & 0.6306\\
    Dev - Task\_A & 0.8535 & \bf{0.8560} & 0.7979\\
    Dev - Task\_B & \bf{0.7387} & 0.7347 & 0.6784\\
    Train-Task\_A & 0.8983 & \bf{0.8990} & 0.8186\\
    Train-Task\_B & \bf{0.7679} & 0.7628 & 0.6704\\
    \hline
\end{tabular}
\caption{Experiments configurations for round 3 and corresponding results compared to best configurations from round 2.}
\label{tbl:exp_round3}
\end{table}

\begin{figure*}[ht]
    \centering
    \includegraphics[width=\textwidth]{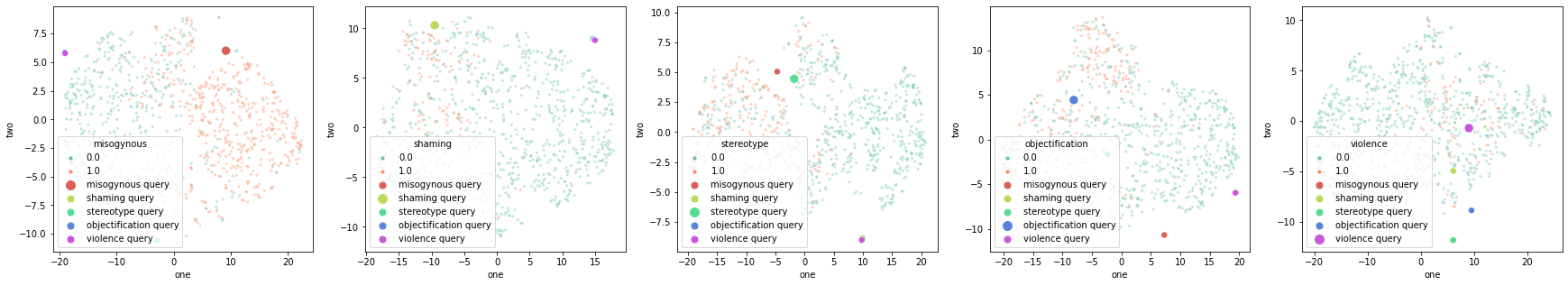}
    \caption{t-SNE projections of transformer decoder output per class (small dots) and the corresponding class query embedding (the biggest dot).}
    \label{fig:embds}
\end{figure*}

\begin{figure*}[ht]
    \begin{subfigure}[t]{\textwidth}
        \centering
        \includegraphics[width=\linewidth]{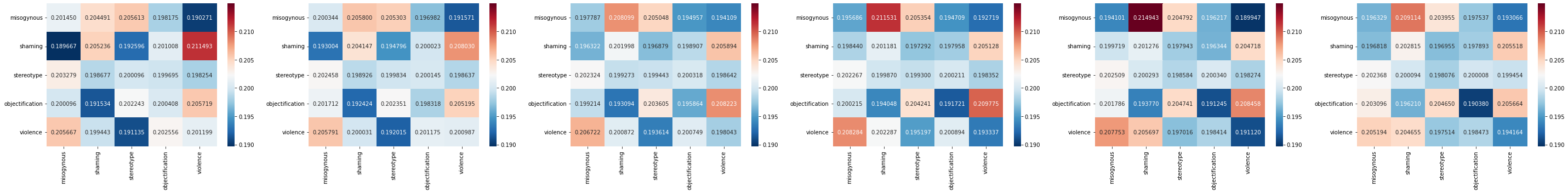}
        \caption{Transformer decoder average self-attention weights}
        \label{fig:decoder_self_attn}
    \end{subfigure}
    \begin{subfigure}[t]{\textwidth}
        \centering
        \includegraphics[width=\linewidth]{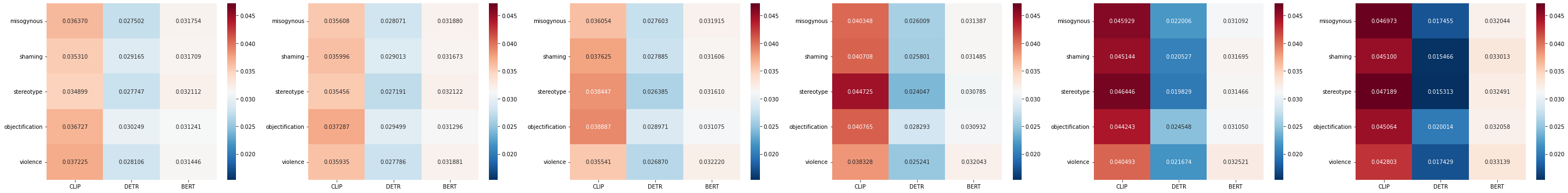}
        \caption{Transformer decoder average source \(\times\) target cross-attention weights. Source weights are aggregated per input encoder.}
        \label{fig:encoder_decoder_cross_attn}
    \end{subfigure}
    \caption{Attention weights visualization per transformer decoder layer. First input layer is on the left while last output layer is on the right.}
    \label{fig:attention_viz}
\end{figure*}

\subsubsection{Round 4}\label{sec:abl_round4}
At this round, we test the significance of the additional training data from Facebook's hateful meme dataset. The configurations and corresponding results are summarized in table \ref{tbl:exp_round4}. This time we train for more \(15\) epochs (i.e. total \(30\) epochs). We can notice a significant improvement when using more training data from the external dataset.

\begin{table}[ht]
\centering
\small
\begin{tabular}{l|l|l}
    \hline
    Experiment & 10 & 30 \\ 
    \hline
    \hline
    Xformer Enc & \multicolumn{2}{c}{Multi} \\
    Pooling & \multicolumn{2}{c}{No} \\
    Proj Algn & \multicolumn{2}{c}{Yes} \\
    Contrastive & \multicolumn{2}{c}{No} \\
    Multi-task & No & Yes \\
    Backbones & \multicolumn{2}{c}{CLIP and DETR} \\
    \hline
    Score \\
    \hline
    Test - Task\_A & 0.7504 & \bf{0.7609}\\
    Test - Task\_B & 0.6798 & \bf{0.6958}\\
    Dev - Task\_A & \bf{0.8535} & 0.8502\\
    Dev - Task\_B & 0.7387 & \bf{0.7429}\\
    Train-Task\_A & 0.8983 & \bf{0.9127}\\
    Train-Task\_B & 0.7679 & \bf{0.7815}\\
    \hline
\end{tabular}
\caption{Experiments configurations for round 4 and corresponding results compared to best configurations from round 2.}
\label{tbl:exp_round4}
\end{table}

\subsection{Visualizations}

In figure \ref{fig:embds}, we show t-SNE projections of the transformer decoder output per class and the corresponding class learnt input query embedding. We use data from experiment \(30\) in section \ref{sec:abl_round4}. It is interesting that, for all classes, the class learnt query embedding is positioned on the side with denser positive labels. This indicates that the learnt class queries can be thought of as centers of positive labels.

In figure \ref{fig:attention_viz}, we show average attention weights per transformer decoder layer. Figure \ref{fig:decoder_self_attn} illustrates the dependencies between each class and other classes. As shown in the figure, numbers are very close which indicates that each class depends uniformly on other classes. Figure \ref{fig:encoder_decoder_cross_attn} shows the average cross-attention weights between the source and target sequences from the transformer decoder layers and aggregated per input encoder. Clearly, the model pays more attention to CLIP features, and less attention to BERT text features, and the least attention to DETR objects features. This conforms with results from round 3 in section \ref{sec:abl_round3}.

\section{Conclusion}

In this paper, we propose a generic framework for both multi-modal embedding and multi-label binary classification tasks. We combine and use as backbones different architectures achieving state-of-the-art in different or similar tasks on different modalities to capture a wide spectrum of semantic signals from the multi-modal input. By employing a multi-task learning scheme, we are able to use multiple datasets from the same knowledge domain and increase the model's performance. In addition to that, we use multiple objectives to regularize and fine tune the system components. We have carried out experiments to verify our ideas and the results show the significance of some of the ideas. As a future work, we need to do more experiments with different backbone architectures and more datasets. We can also try more objectives and regularizations; for instance, we can use our observation from figure \ref{fig:embds} to make the decoder output for a positive instance closer to the corresponding class query embedding.



\bibliography{acl_latex}
\bibliographystyle{acl_natbib}

\appendix

\section{Experiments scores distributions}
\label{sec:appendix}

\begin{figure*}[ht]

\begin{tabular}{ccccc}
  \hline
  & \multicolumn{2}{c}{Task\_A} & \multicolumn{2}{c}{Task\_B} \\ 
  \hline
  & Tukey's plot & Box plot & Tukey's plot & Box plot \\
  \hline
    \rotatebox{90}{\quad Round 1} &
    \begin{subfigure}[t]{0.21\textwidth}
        \centering
        \includegraphics[width=\linewidth]{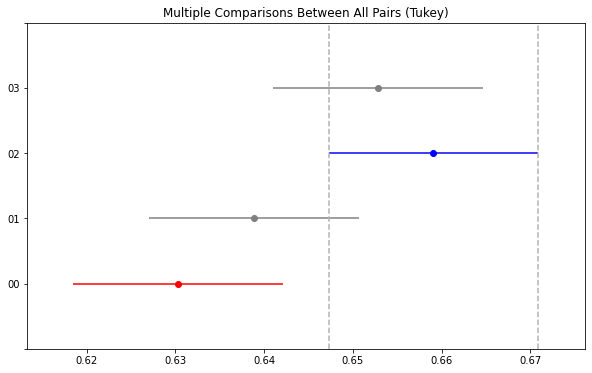}
        \caption{}
        \label{fig:exp_1_a_tucky}
    \end{subfigure}
    &
    \begin{subfigure}[t]{0.21\textwidth}
        \centering
        \includegraphics[width=\linewidth]{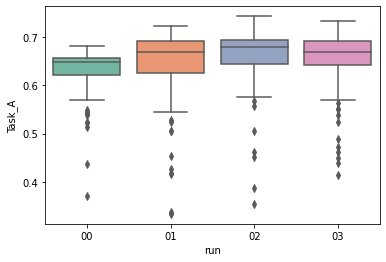}
        \caption{}
        \label{fig:exp_1_a_box}
    \end{subfigure}
    &
    \begin{subfigure}[t]{0.21\textwidth}
        \centering
        \includegraphics[width=\linewidth]{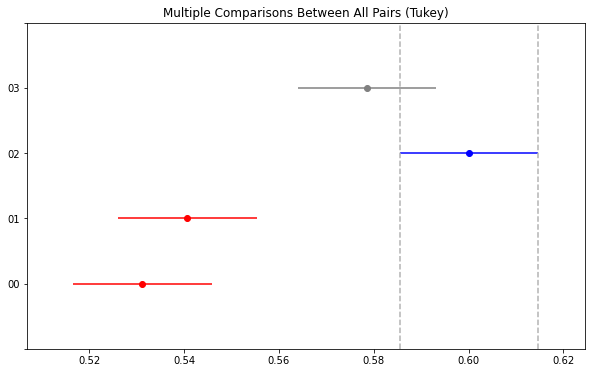}
        \caption{}
        \label{fig:exp_1_b_tucky}
    \end{subfigure}
    &
    \begin{subfigure}[t]{0.21\textwidth}
        \centering
        \includegraphics[width=\linewidth]{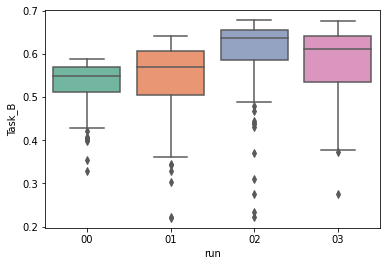}
        \caption{}
        \label{fig:exp_1_b_box}
    \end{subfigure} \\
  \hline
  
   \rotatebox{90}{\quad Round 2} &
    \begin{subfigure}[t]{0.21\textwidth}
        \centering
        \includegraphics[width=\linewidth]{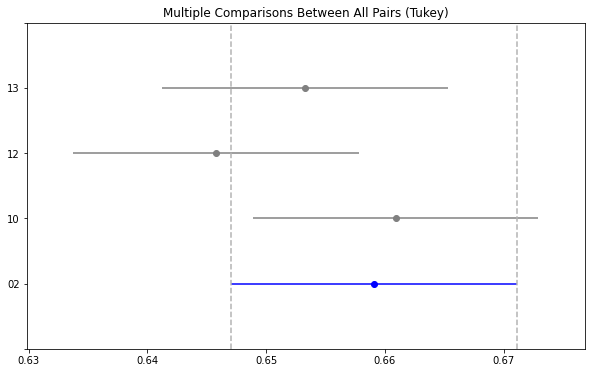}
        \caption{}
        \label{fig:exp_2_a_tucky}
    \end{subfigure}
    &
    \begin{subfigure}[t]{0.21\textwidth}
        \centering
        \includegraphics[width=\linewidth]{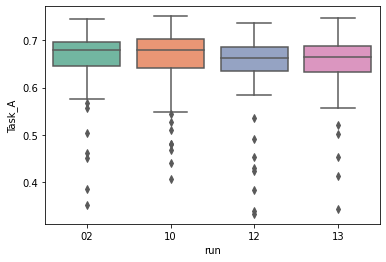}
        \caption{}
        \label{fig:exp_2_a_box}
    \end{subfigure}
    &
    \begin{subfigure}[t]{0.21\textwidth}
        \centering
        \includegraphics[width=\linewidth]{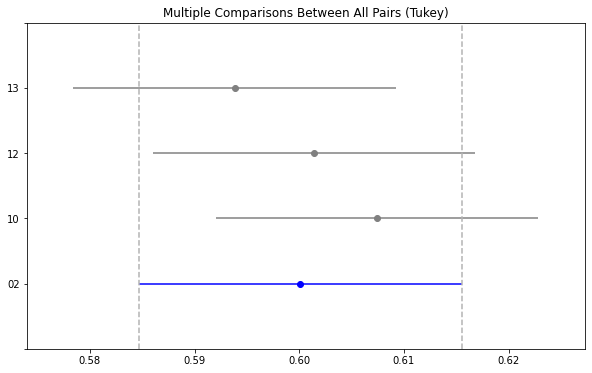}
        \caption{}
        \label{fig:exp_2_b_tucky}
    \end{subfigure}
    &
    \begin{subfigure}[t]{0.21\textwidth}
        \centering
        \includegraphics[width=\linewidth]{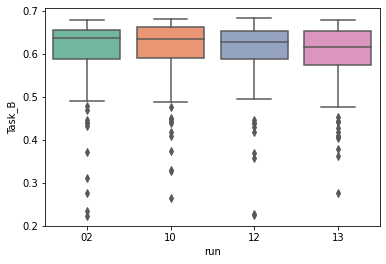}
        \caption{}
        \label{fig:exp_2_b_box}
    \end{subfigure} \\
  \hline
  
   \rotatebox{90}{\quad Round 3} &
    \begin{subfigure}[t]{0.21\textwidth}
        \centering
        \includegraphics[width=\linewidth]{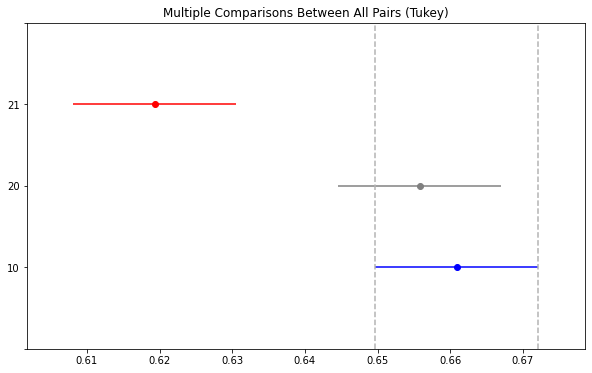}
        \caption{}
        \label{fig:exp_3_a_tucky}
    \end{subfigure}
    &
    \begin{subfigure}[t]{0.21\textwidth}
        \centering
        \includegraphics[width=\linewidth]{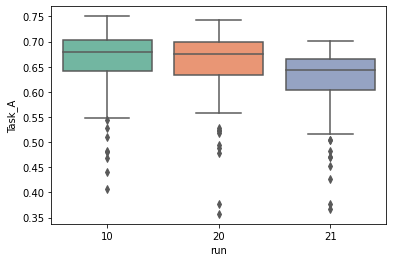}
        \caption{}
        \label{fig:exp_3_a_box}
    \end{subfigure}
    &
    \begin{subfigure}[t]{0.21\textwidth}
        \centering
        \includegraphics[width=\linewidth]{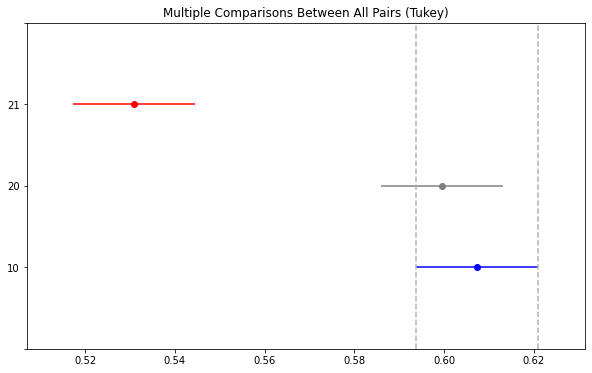}
        \caption{}
        \label{fig:exp_3_b_tucky}
    \end{subfigure}
    &
    \begin{subfigure}[t]{0.21\textwidth}
        \centering
        \includegraphics[width=\linewidth]{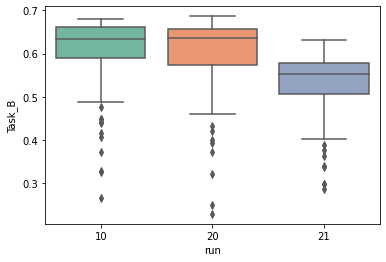}
        \caption{}
        \label{fig:exp_3_b_box}
    \end{subfigure} \\
  \hline
  
    \rotatebox{90}{\quad Round 4} &
    \begin{subfigure}[t]{0.21\textwidth}
        \centering
        \includegraphics[width=\linewidth]{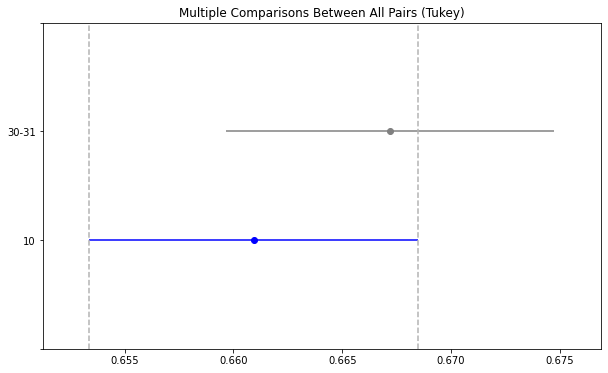}
        \caption{}
        \label{fig:exp_4_a_tucky}
    \end{subfigure}
    &
    \begin{subfigure}[t]{0.21\textwidth}
        \centering
        \includegraphics[width=\linewidth]{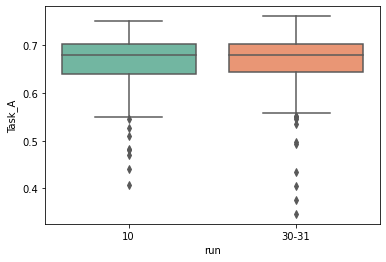}
        \caption{}
        \label{fig:exp_4_a_box}
    \end{subfigure}
    &
    \begin{subfigure}[t]{0.21\textwidth}
        \centering
        \includegraphics[width=\linewidth]{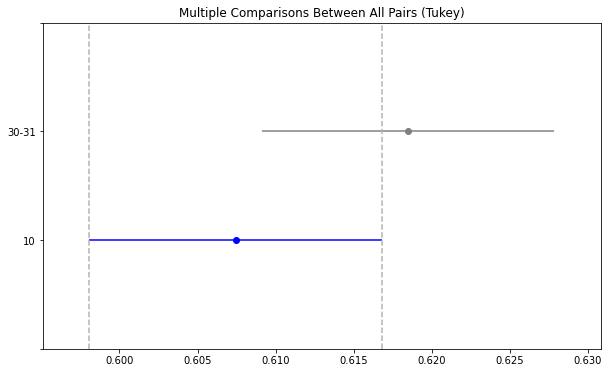}
        \caption{}
        \label{fig:exp_4_b_tucky}
    \end{subfigure}
    &
    \begin{subfigure}[t]{0.21\textwidth}
        \centering
        \includegraphics[width=\linewidth]{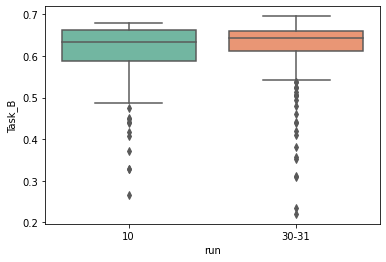}
        \caption{}
        \label{fig:exp_4_b_box}
    \end{subfigure} \\
    
  \hline
\end{tabular}

    \caption{Performance comparison on test split for Task\_A and Task\_B collected from evaluation phases during model training epochs for different experiments described in \ref{sec:ablations}. Tukey's plots visualize a universal confidence interval of scores mean on x-axis for each run on y-axis, any two runs can be compared for significance by looking for overlap. Box plots summarize the distribution of scores on y-axis for each run on x-axis.}
    \label{fig:exp_dist_test}
\end{figure*}

\end{document}